\pdfoutput=1
\documentclass[11pt]{article}

\usepackage[final]{acl}

\usepackage{times}
\usepackage{latexsym}
\usepackage{enumitem}
 \usepackage{amsmath}
\usepackage[T1]{fontenc}

\usepackage[utf8]{inputenc}

\usepackage{microtype}

\usepackage{inconsolata}

\usepackage{graphicx}

\usepackage[table]{xcolor}
\usepackage{booktabs}   
\usepackage{colortbl}   
\usepackage{array}
\usepackage{caption}
\usepackage{subcaption} 
\usepackage{adjustbox}
\usepackage{float} 
\definecolor{stronggreen}{RGB}{0,128,0}       
\definecolor{moderategreen}{RGB}{144,238,144} 
\definecolor{lightorange}{RGB}{255,200,0}     

\title{\textsc{CLIN}: an Objective Framework for Evaluating Creativity \\in Short Persian Literary Text}

\author{
 \textbf{Mohammad Reza Modarres*\textsuperscript{1,2}},
 \textbf{Armin Tourajmehr*\textsuperscript{2}},\\
 \textbf{Yadollah Yaghoobzadeh\textsuperscript{\ensuremath{\dagger}}\textsuperscript{1,2}},
 \textbf{Mohammad Taher Pilehvar\textsuperscript{\ensuremath{\dagger}}\textsuperscript{3}}
\\
 \textsuperscript{1}School of Electrical and Computer Engineering, \\
 College of Engineering, University of Tehran, Tehran, Iran\\
 \textsuperscript{2}Tehran Institute for Advanced Studies, Khatam University, Iran \\
 \textsuperscript{3}Cardiff University, United Kingdom
\\
 \small{
   \textbf{Correspondence:} 
   \href{mailto:mr.modarres@ut.ac.ir}{mr.modarres@ut.ac.ir}
 }
}

\begin{document}
\maketitle

\begingroup
  \renewcommand{\thefootnote}{*,\ensuremath{\dagger}}
  \footnotetext{Equal contribution, equal supervision}

\endgroup

\begin{abstract}

Evaluating creativity in large language model (LLM) outputs remains challenging because creativity is multidimensional and human-centered. We examine how reliably LLMs evaluate short literary text in Persian, a low-resource language, across multiple evaluation strategies and prompt formulations. We find that LLM--human agreement varies substantially across dimensions: alignment is stronger for structured TTCT-derived properties such as \emph{Originality}, \emph{Fluency}, and \emph{Elaboration}, but considerably weaker for more subjective dimensions, particularly \emph{Emotion} and \emph{Attractiveness}. Judgments are also sensitive to prompt formulation, while few-shot prompting, ensembling, and multi-agent debate provide no consistent improvement. Motivated by this dimension-dependent behavior, we investigate whether structured creativity dimensions can instead be approximated using simple, interpretable proxy metrics. We introduce \textsc{CLIN}, which evaluates three TTCT-derived dimensions separately using topic-aware novelty for \emph{Originality}, contextual lexical clustering for \emph{Fluency}, and lexical diversity for \emph{Elaboration}. These proxies achieve human alignment comparable to or better than the strongest zero-shot LLM judge in our setting while requiring substantially lower evaluation cost.\footnote{Code, data, and additional materials are available at \url{https://smrmodares.github.io/CLiN/}.}
\end{abstract}

\section{Introduction}

With the rapid advancement of LLMs, creative text generation has received growing attention across applications ranging from story and poetry generation to assistive writing systems \citep{wen2023grove,zhou2023recurrentgpt,elzohbi2023creative,yuan2022wordcraft}. Alongside progress in generation, evaluating creativity has itself become an active research problem, with approaches including psychometric evaluation, LLM-based judging, reference-based comparison, learned evaluators, and automated creativity measures \citep{chakrabarty2023art,li2025automated,ismayilzada-etal-2025-creative,fein-etal-2026-litbench,lu-etal-2026-rethinking}. Despite this progress, substantial disagreement remains over which signals reflect human creativity judgments and when automated evaluations can be trusted.

Creativity is a human-centered and multidimensional construct. Psychometric frameworks such as the Alternative Uses Test \cite[AUT]{guilford1967creativity} and Torrance Tests of Creative Thinking \cite[TTCT]{torrance1966ttct} address this complexity by separating creativity into complementary dimensions, an approach increasingly adopted in NLP evaluation \citep{chakrabarty2023art,tourajmehr-etal-2025-evaluating}. However, evidence on the reliability of LLM creativity evaluators is mixed: model--human agreement varies across criteria \citep{app15062971,lyu2026largelanguagemodelshumans}, reference-based evaluation can improve alignment \citep{li2025automated}, specialized evaluators can outperform off-the-shelf LLM judges \citep{fein-etal-2026-litbench}, and judgments can be sensitive to relatively minor prompt changes \citep{lu-etal-2026-rethinking}. We therefore ask how reliably LLM judges recover human assessments across different creativity dimensions and evaluation protocols, and whether structured dimensions can instead be approximated using simple, interpretable measurements.

To address these questions, we conduct a systematic evaluation of LLM-based creativity judgments across multiple experimental settings, including zero-shot evaluation, reference-based comparison, few-shot prompting, ensemble voting, multi-agent debate, and prompt variation. We focus on short literary text written in Persian, a low-resource language that poses additional challenges for both creative generation and evaluation.

Our results show that LLMs achieve moderate alignment with human judgments on relatively structured dimensions such as \emph{Originality}, \emph{Fluency}, and \emph{Elaboration}, but considerably weaker alignment on more subjective dimensions, particularly \emph{Creativity}, \emph{Attractiveness}, and \emph{Emotion}. More elaborate judging procedures provide no consistent improvement, and LLM-based evaluations remain sensitive to prompt formulation.

These results motivate a different evaluation strategy for the structured dimensions. We introduce \textsc{CLIN} (Creativity as Lexical Ideas, Novelty, and N-grams), an objective and interpretable framework for the TTCT-derived dimensions considered in this work. \textsc{CLIN} evaluates \emph{Originality} through global and topic-relative novelty, \emph{Fluency} through contextual lexical clustering, and \emph{Elaboration} through lexical diversity. Each dimension is evaluated separately rather than collapsed into a single creativity score. Our analysis further shows that these simple, objective, and interpretable proxies can perform comparably to or better than the strongest zero-shot LLM judge in our setting, while requiring substantially lower evaluation cost.

In summary, our contributions are as follows:
\begin{itemize}
    \item A systematic evaluation of LLMs as creativity judges in short Persian literary text across multiple evaluation strategies and prompt formulations.
    
    \item Empirical evidence that LLM-based judgments exhibit limited alignment with human evaluations, particularly for subjective dimensions such as emotion and attractiveness, and remain sensitive to changes in prompt formulation.
    
    \item The introduction of \textsc{CLIN}, an objective and interpretable proxy-based framework for evaluating \emph{Originality}, \emph{Fluency}, and \emph{Elaboration}, whose simple lexical and statistical measures achieve alignment with human judgments comparable to or better than the strongest zero-shot LLM judge at substantially lower evaluation cost.
    
\end{itemize}

\section{Related Work}
\paragraph{Creativity as a multidimensional construct.}
Creativity is commonly treated as a multidimensional construct rather than a single observable property. Psychometric frameworks such as the Alternative Uses Test \cite[AUT]{guilford1967creativity}, Divergent Association Task \citep[DAT]{olson2021naming}, and Torrance Tests of Creative Thinking \cite[TTCT]{torrance1966ttct} measure complementary aspects of creative behavior, with TTCT distinguishing originality, fluency, flexibility, and elaboration. Recent computational work similarly emphasizes dimensions beyond novelty, including diversity, surprise, quality, and appropriateness \citep{ismayilzada-etal-2025-creative,nakajima-etal-2026-beyond}. Following this view, we evaluate TTCT-derived dimensions separately using dedicated, interpretable proxies.

\paragraph{LLMs as creativity evaluators.}
LLMs are increasingly used to evaluate creative outputs, but their agreement with humans varies across settings. Prior work reports weak agreement on several creativity dimensions \citep{chakrabarty2023art}, while other studies find stronger consistency but limitations in culturally or contextually dependent judgments \citep{app15062971}. Reference-based evaluation can improve alignment \citep{li2025automated}, and LitBench shows that specialized reward models can outperform zero-shot LLM judges \citep{fein-etal-2026-litbench}. Recent studies further show substantial disagreement among creativity metrics, sensitivity of LLM judgments to prompt variation \citep{lu-etal-2026-rethinking}, and criterion-dependent LLM--human agreement \citep{lyu2026largelanguagemodelshumans}. Our study complements this work by evaluating the same texts and human-rated dimensions across zero-shot, reference-based, few-shot, ensemble, multi-agent, and prompt-variation settings.

\paragraph{Automated creativity measures.}
A parallel line of work seeks scalable alternatives to direct human or LLM judgment. CS4 measures story creativity through writing constraints \citep{atmakuru2024cs4}, PACE evaluates associative creativity through semantic associations \citep{qiu-hu-2025-deep}, and recent work has explored inexpensive metrics such as perplexity and corpus overlap \citep{lu-etal-2026-rethinking} as well as domain-general evaluation using semantic entropy and multi-agent methods \citep{sen-etal-2026-automated}.

\textsc{CLIN} differs from these approaches in both granularity and objective. Rather than learning a holistic judge or constructing a task-general measure of model creativity, \textsc{CLIN} evaluates individual literary products along predefined human-rated dimensions using separate, transparent proxies: global and topic-relative novelty for originality, contextual lexical clustering for fluency, and lexical diversity for elaboration. Its goal is to determine whether such low-cost, dimension-specific measurements can recover human judgments without requiring a generative evaluator, learned reward model, or reference exemplar.

\section{Creativity Evaluation Test}\label{sec:introduced_metrics}

To evaluate creativity in literary text, we build on established psychometric approaches. We adopt the TTCT as a structured foundation, following recent work such as \citet{chakrabarty2023art}, who introduced the Torrance Test of Creative Writing (TTCW) for assessing AI-generated stories. Their findings show that LLM-generated stories perform substantially worse than professional human-written stories across TTCW-based creativity assessments, indicating a persistent gap in creative writing performance.

Inspired by this line of work, we assess creativity using three TTCT-derived dimensions: \textit{Originality}, \textit{Fluency}, and \textit{Elaboration}. The \textit{Flexibility} dimension is excluded, as prior studies have shown it to be highly correlated with Fluency and thus redundant in practice \cite{weiss2022flexibility, kackenmester2019openness}. Specifically, Originality measures the novelty and non-cliché nature of a text; Fluency captures the diversity of ideas used to convey the topic; and Elaboration evaluates the depth of development and lexical richness. These dimensions provide a structured and interpretable basis for systematic evaluation.

However, TTCT-style metrics alone do not fully capture the holistic and subjective nature of creativity. Prior work by \citet{marco-etal-2024-pron} proposes an alternative rubric grounded in Boden’s notion of creativity,\footnote{Creativity is the ability to
come up with ideas that are new, surprising and
valuable \citep{boden2004creative}.} incorporating dimensions such as Creativity, Attractiveness, Originality, Anthology Potential, and Own Voice to evaluate short stories. While their framework is well-suited to longer narrative texts, our setting differs substantially: we focus on short literary text, often consisting of a single sentence, where several of these dimensions are less applicable or difficult to assess reliably.

We therefore adopt a complementary approach. Retaining the structured TTCT dimensions as a backbone, we incorporate a subset of subjective dimensions inspired by prior work, specifically \textit{Creativity} and \textit{Attractiveness}, that are compatible with our shorter text format. The Creativity dimension captures overall inventiveness, while Attractiveness reflects aesthetic appeal and reader engagement. In addition, motivated by prior evidence that emotional expression is a core component of creative writing \cite{Kaufman_Sternberg_2019}, as well as findings from \citet{chakrabarty2023art} highlighting LLMs’ limitations in modeling emotional depth, we introduce an \textit{Emotion} dimension to assess the intensity and vividness of affective content.

Importantly, we distinguish between structured and holistic evaluation. The TTCT-derived dimensions provide a decomposition of creativity into interpretable components, while the overall \textit{Creativity} score captures a holistic human judgment of creativity as an integrated construct and is not defined as a simple aggregation of the other dimensions. Our goal is twofold: first, to examine how individual TTCT dimensions relate to holistic human perceptions of creativity, and second, to evaluate how well structured, proxy-based metrics align with these subjective judgments. All dimensions are rated following a consistent scoring procedure, in which annotators assign scores based on the rubric definitions provided in Appendix~\ref{tab:appendix_metrics}. Both human annotators and LLM evaluators were provided with the same evaluation questionnaire, given in Appendix~\ref{appendix:prompt}, to ensure that judgments were based on identical criteria and instructions.

\section{Evaluating LLM as Judge}
To investigate the reliability of LLMs as evaluators of creativity, we conduct a series of experiments in which LLMs judge literary text under varying evaluation scenarios and prompt formulations. This section first outlines the experimental setup, including the data and scoring procedures, and then presents each experiment along with its results, highlighting the degree of alignment between LLM-based judgments and human evaluations.

\subsection{Dataset}

Our evaluation dataset consists of 200 short Persian literary texts, including 100 human-authored texts and 100 GPT-3.5-turbo–generated \cite{openai2023chatgpt} texts from the CPers dataset \citep{tourajmehr-etal-2025-evaluating}. The texts cover five thematic categories: hope, despair, longing, love, and friendship. Including both human-authored and model-generated texts allows us to investigate whether LLM-based evaluation varies for texts originating from different sources.

Five human annotators independently evaluated all texts using a 3-point rating scale (1--3), following the scoring approach of \citet{app15062971}. 
Each text was rated along six dimensions as defined in Section~\ref{sec:introduced_metrics}: \textit{Originality}, \textit{Fluency}, \textit{Elaboration}, \textit{Creativity}, \textit{Attractiveness}, and \textit{Emotion}. 
Final scores for each dimension are obtained via averaging across annotators. The correlations between individual annotators and the aggregated scores are provided in {Appendix~\ref{appendix:annotator_singlecol}}.

\begin{figure}[t!]
    \centering
    \includegraphics[width=\columnwidth]{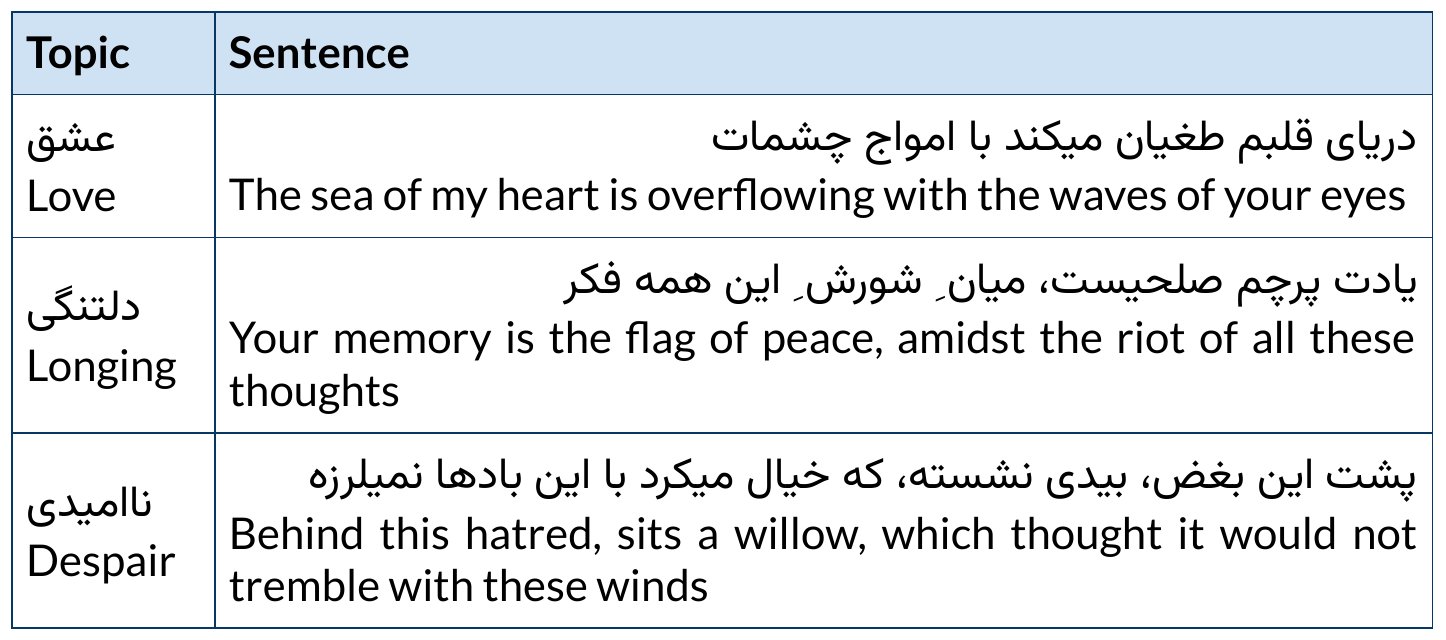}
    \caption{Sample Persian instances from the dataset.}
    \label{fig:sample}
\end{figure}

Overall, this process yields a dataset of 200 texts with human average scores, serving as the reference signal for evaluating the alignment of LLM-based judgments across creativity dimensions. Figure \ref{fig:sample} showcases some examples of the dataset.

\begin{figure}[t]
    \centering
    \includegraphics[
        width=0.90\columnwidth
    ]{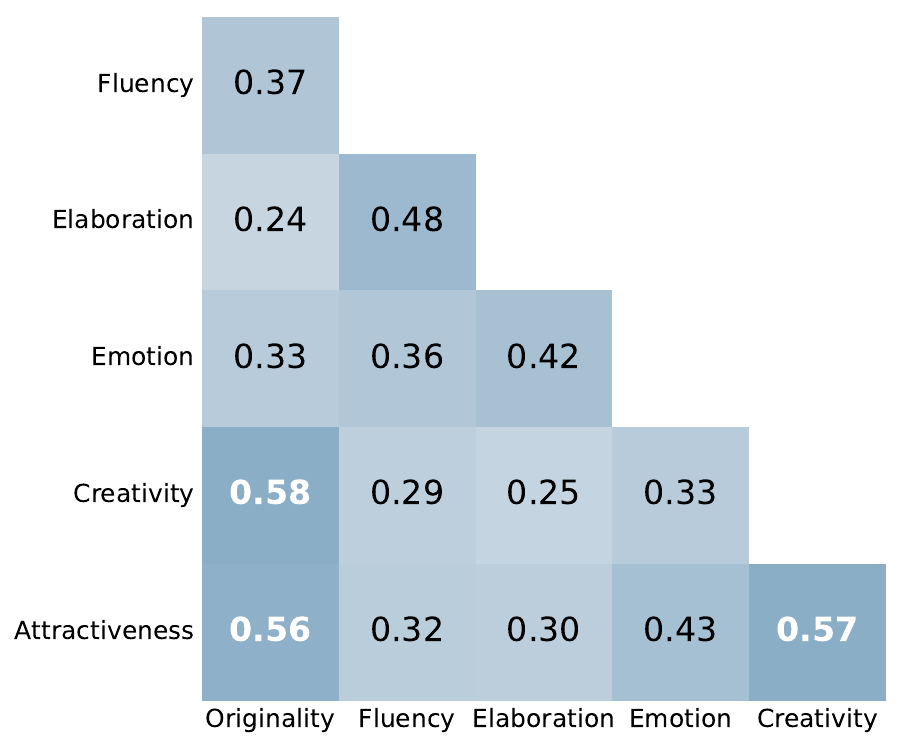}
    \caption{Pairwise Spearman correlations among the six human-rated evaluation dimensions. Values represent Spearman's rank correlation coefficient ($\rho$) between each pair of dimensions. Only the lower triangle is shown for clarity.}
    \label{fig:spearman_comparison}
\end{figure}

\subsection{Relationships Among Human-Rated Dimensions}

To further characterize the human annotations, we examine how the six evaluation dimensions relate to one another. We compute pairwise Spearman rank correlations across all 200 texts (Figure~\ref{fig:spearman_comparison}). This analysis provides a descriptive view of the annotation space, showing which dimensions tend to vary together and which exhibit weaker associations.

The strongest associations involve Originality, Creativity, and Attractiveness. Originality correlates most strongly with Creativity ($\rho=0.58$) and Attractiveness ($\rho=0.56$), while Creativity and Attractiveness are similarly associated ($\rho=0.57$). Thus, within this dataset, holistic creativity judgments tend to co-vary most strongly with perceived originality and aesthetic appeal.

A different pattern emerges for the remaining dimensions. Fluency and Elaboration show a moderate positive association ($\rho=0.48$), while Emotion is moderately associated with both Elaboration ($\rho=0.42$) and Attractiveness ($\rho=0.43$). In contrast, Creativity is more weakly associated with Fluency ($\rho=0.29$), Elaboration ($\rho=0.25$), and Emotion ($\rho=0.33$). These differences indicate that the dimensions are not equally associated with holistic creativity judgments.

Taken together, the human-rated dimensions are related but not interchangeable. The stronger associations among Originality, Creativity, and Attractiveness reveal one pattern of co-variation, while Fluency, Elaboration, and Emotion exhibit different relationships within the annotation space. These findings are descriptive and should not be interpreted as evidence that the dimensions correspond to distinct latent constructs or causal relationships.

\begin{figure*}[t!]
    \centering
    \begin{subfigure}[b]{0.48\textwidth}
        \centering
        \includegraphics[width=\textwidth]{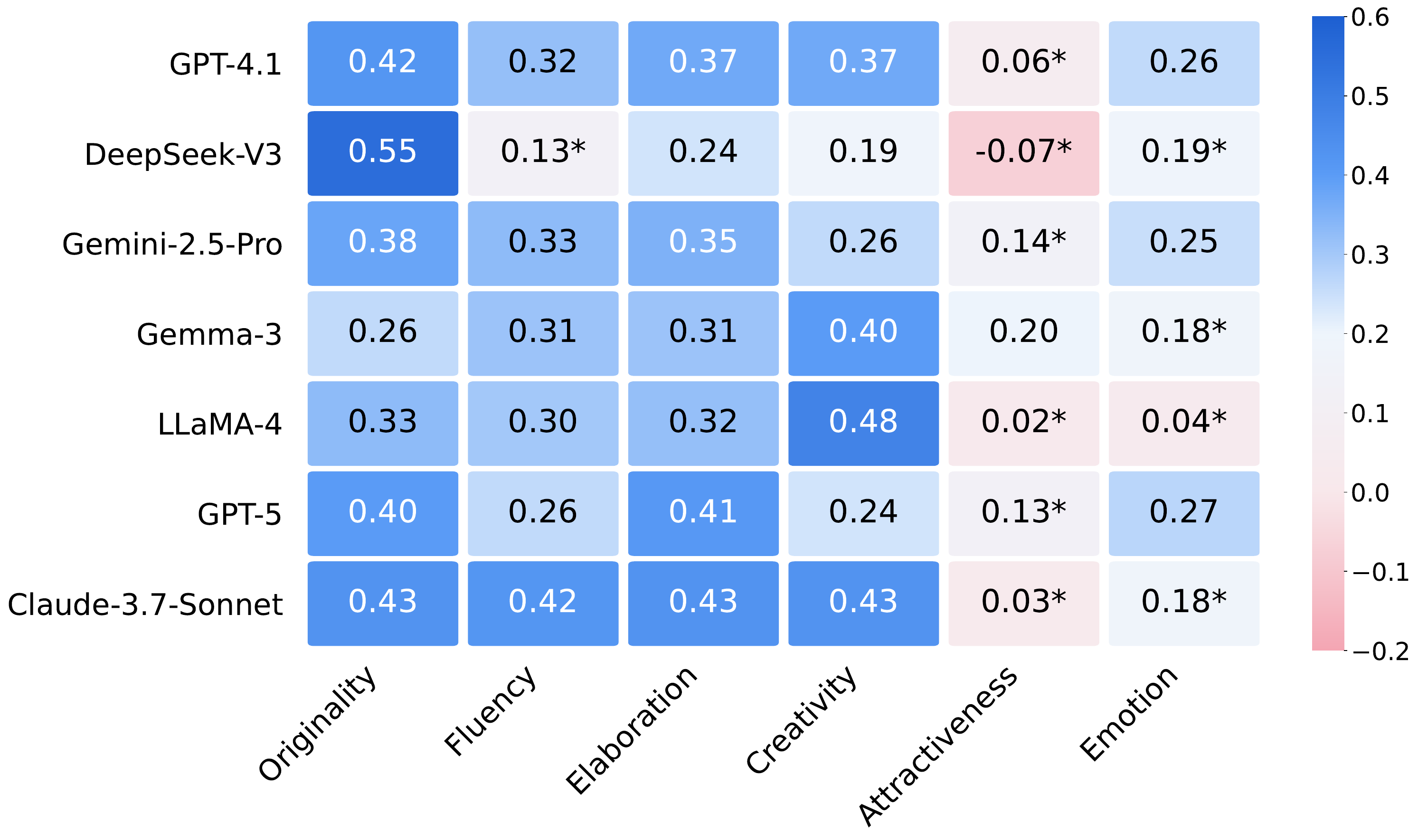}
        \caption{Human-authored text}
        \label{fig:human_texts_corr}
    \end{subfigure}
    \begin{subfigure}[b]{0.48\textwidth}
        \centering
        \includegraphics[width=\textwidth]{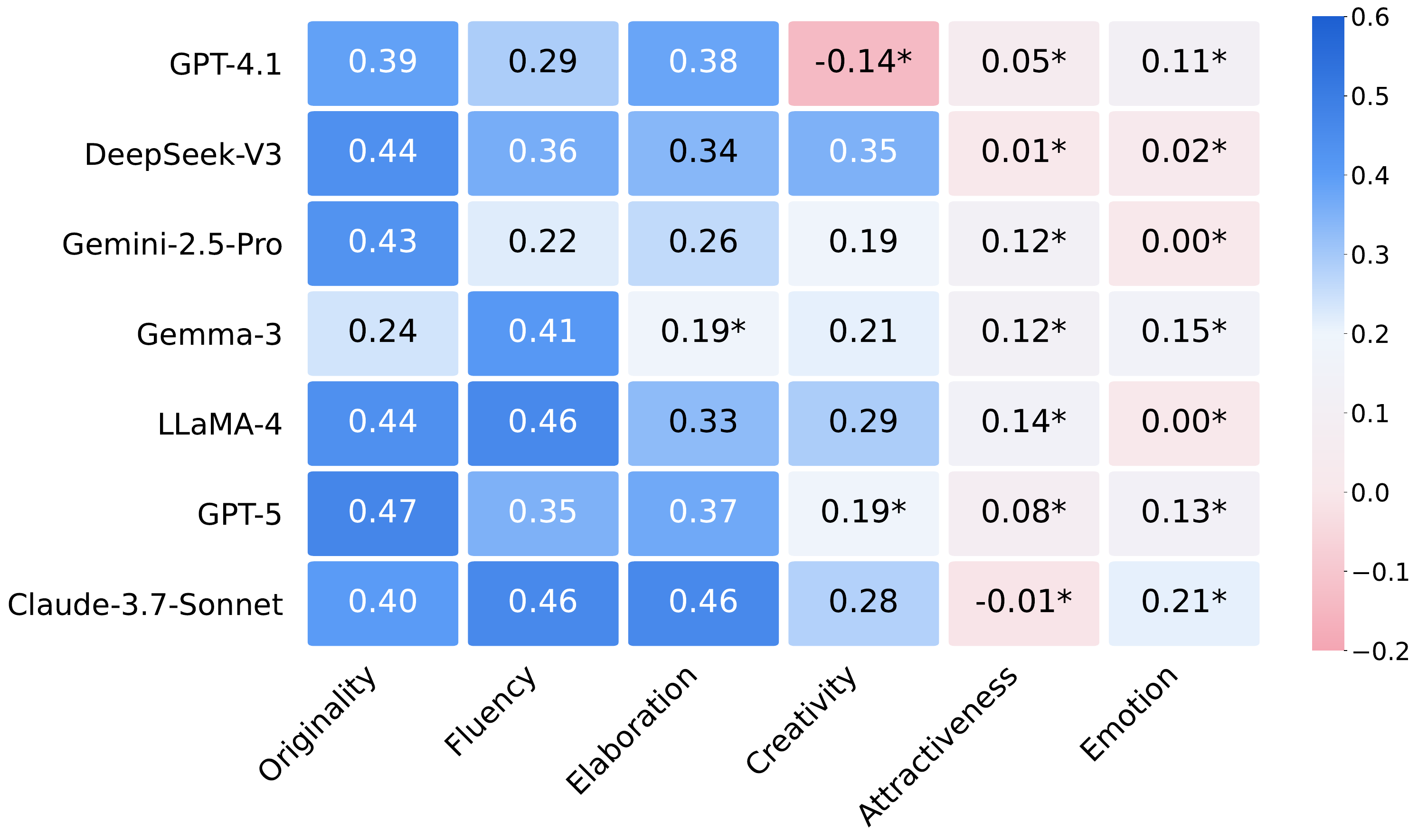}
        \caption{Model-generated text}
        \label{fig:model_generated_corr}
    \end{subfigure}
    \hfill
    \caption{Spearman correlation between model-assigned scores and human ratings across different dimensions for (a) human-authored text and (b) model-generated text. Values marked with $*$ are \textit{not} statistically significant.}
    \label{fig:combined_corr}
\end{figure*}

\subsection{Experiments}

The reliability of LLMs as creativity judges is evaluated across a range of settings designed to test whether increasing evaluation sophistication improves alignment with human judgments. Four experimental configurations are considered, differing in the number of models and their prompting or coordination strategies, ranging from single-model zero-shot evaluation to ensemble methods and multi-agent debate frameworks. This setup enables a systematic examination of whether additional context, interaction, or aggregation improves the reliability of creativity judgments.

Experiments are conducted using seven models: GPT-4.1 \cite{openai2025gpt41}, GPT-5 \cite{openai2025gpt5}, Claude 3.7 Sonnet \cite{anthropic2025claude37}, LLaMA-4 \cite{meta2025llama4}, Gemini 2.5 Pro \cite{google2025gemini25pro}, Gemma-3 \cite{gemma2025instruct}, and DeepSeek-V3 \cite{deepseek2025v30324}. Each model evaluates the same set of texts across multiple creativity dimensions, and the resulting scores are compared against human annotations.

Agreement between model outputs and human judgments is measured using Spearman’s rank correlation for each creativity dimension, providing a standard measure of consistency between evaluators. Spearman’s correlation is appropriate because our goal is to evaluate whether the relative ordering of model-generated scores aligns with the ordering induced by human annotators, rather than comparing exact score values. The four experimental settings progressively introduce increasing levels of prompting structure and model coordination; each is described in detail below, along with the corresponding results.

\subsubsection{Single-judge Evaluation}

Each model is first evaluated independently by scoring literary texts across the six proposed dimensions using the same prompts as human annotators (Appendix~\ref{appendix:prompt}). Figure~\ref{fig:human_texts_corr} shows correlations between model-assigned scores and human ratings for human-authored texts. Among the evaluated models, DeepSeek-V3 achieves the highest agreement with human judgments on \textit{originality}, LLaMA-4 performs best on \textit{creativity}, and Claude~3.7~Sonnet performs best on \textit{fluency} and \textit{elaboration}. Considering performance across the three TTCT-derived dimensions as a whole, Claude~3.7~Sonnet is the strongest overall zero-shot judge.

Despite these relative strengths, no model exhibits satisfactory alignment with human raters on the more subjective dimensions of creativity, namely \textit{attractiveness} and \textit{emotion}. Correlations for these dimensions are consistently low, and for \textit{emotion}, most fail to reach statistical significance.


Figure~\ref{fig:model_generated_corr} presents the corresponding analysis for model-generated texts. Claude~3.7~Sonnet achieves the strongest average alignment across the three TTCT-derived dimensions, \textit{originality}, \textit{fluency}, and \textit{elaboration}, although the best-performing model varies by individual dimension.

To further examine whether evaluation behavior differs according to the source of the text, we average the model--human correlations across the seven evaluators for each dimension. As shown in Table~\ref{tab:source_comparison}, alignment is relatively stable across human-authored and model-generated texts for the three TTCT-derived dimensions, with fluency showing the largest increase on model-generated texts. In contrast, agreement with human judgments decreases considerably for overall creativity and emotion when evaluating model-generated texts, while attractiveness remains essentially unchanged. This pattern suggests that LLM-based evaluation is relatively robust to text source for the more structured TTCT-derived dimensions, whereas judgments of holistic creativity and emotion are more sensitive to whether the evaluated text is human- or model-generated.

\begin{table}[t]
\centering
\small
\begin{tabular}{lccc}
\toprule
\textbf{Dimension} & \textbf{Human} & \textbf{Model-generated} & $\boldsymbol{\Delta}$ \\
\midrule
Originality     & 0.396 & 0.401 & $+$0.006 \\
Fluency         & 0.296 & 0.364 & $+$0.069 \\
Elaboration     & 0.347 & 0.333 & $-$0.014 \\
\midrule
Emotion         & 0.196 & 0.089 & $-$0.107 \\
Creativity      & 0.339 & 0.196 & $-$0.143 \\
Attractiveness  & 0.073 & 0.073 &  ~~~0.000 \\
\bottomrule
\end{tabular}
\caption{Average Spearman correlation between LLM-based evaluations and human ratings for human-authored and model-generated texts. Values are averaged across the seven LLM evaluators in Figure~\ref{fig:combined_corr}. $\Delta$ denotes the difference between model-generated and human-authored texts.}
\label{tab:source_comparison}
\end{table}

We further investigate whether providing additional context improves evaluation reliability by introducing a reference-based comparison setting. Instead of assigning absolute scores, models are asked to compare a candidate text against a reference and select which is better for a given dimension. At a high level, this formulation aims to reduce calibration issues by reframing evaluation as a relative judgment task.
However, we find that model outputs are highly sensitive to the choice of reference, and no single reference consistently improves performance across dimensions or instances.

Overall, this setting yields less reliable results than the single-judge setup. Full experimental details and results for the top-performing models (Claude, DeepSeek, and GPT-5) across all six dimensions are provided in Appendix~\ref{appendix:comparative}.

\subsubsection{Few-shot Prompt Evaluation}

The impact of few-shot prompting on the reliability of LLM-based creativity judgments is evaluated by providing models with human-annotated examples that illustrate the expected scoring behavior. Due to cost considerations, we first conducted pilot experiments across multiple models. In these experiments, few-shot prompting generally reduced correlation with human judgments across the evaluated models. Among the few-shot configurations, GPT-4.1 achieved the strongest performance and was therefore selected for the analysis reported here. Consistent with our goal of reporting the strongest achievable performance under practical constraints, subsequent experiments focus on the best-performing configuration in each setting.

Three few-shot configurations are considered: (1) examples with the highest human-assigned scores, (2) examples with the lowest scores, and (3) a mixed set containing high-, medium-, and low-scored examples. Among these, the mixed configuration yields the strongest performance and is therefore reported in Table~\ref{tab:human_model_correlation}.


Across all configurations, model-assigned scores remain largely unchanged relative to the zero-shot setting, and in several cases show lower alignment with human judgments. These results suggest that few-shot prompting does not improve LLM-based creativity evaluation and may even degrade it.

\begin{table}[]
\centering
\setlength{\tabcolsep}{4pt}
\renewcommand{\arraystretch}{1.15}
\scalebox{0.8}{
\begin{tabular}{lcccccc}
\toprule
\textbf{Strategies} & \textbf{Org.} & \textbf{Flu.} & \textbf{Ela.} & \textbf{Emo.} & \textbf{Crea.} & \textbf{Attr.} \\
\midrule
Few-shot & 0.34 & 0.35 & \textbf{0.50} & 0.29 & 0.34 & 0.18$^*$ \\
Majority weak & 0.36 & {0.40} & 0.28 & ~~0.11$^*$ & \textbf{0.42} & 0.03$^*$ \\
Majority strong & 0.41 & \textbf{0.46} & 0.41 & 0.32 & 0.31 & 0.06$^*$ \\
Multi-agent & 0.42 & 0.44 & \textbf{0.45} & 0.34 & 0.37 & 0.12$^*$ \\
\bottomrule
\end{tabular}
}
\caption{Spearman correlation between model-based evaluations and human ratings across evaluation strategies, averaged over human- and model-generated texts. Values marked with $*$ are not statistically significant.}
\label{tab:human_model_correlation}
\end{table}

\subsubsection{Ensemble Evaluation}
An ensemble-based evaluation is conducted by aggregating model judgments via majority voting. Specifically, an ensemble of three of the strongest models (Claude, Gemini, and GPT-5) is evaluated by comparing aggregated scores with human ratings. The same procedure is applied to a second ensemble of comparatively weaker models: DeepSeek, LLaMA-4, and Gemma-3.

As shown in Table~\ref{tab:human_model_correlation}, in both cases, correlations are comparable to those of the best individually performing model on each dimension, with neither ensemble improving alignment with human judgments. Performance remains relatively strong for \textit{originality}, \textit{fluency}, and \textit{elaboration}, but consistently weak for \textit{emotion}, \textit{creativity}, and \textit{attractiveness}. These results indicate that simple ensemble strategies are insufficient to overcome the limitations of LLMs in evaluating subjective aspects of creativity.

\subsubsection{Multi-agent Debate Evaluation}
A multi-agent evaluation framework is implemented using Gemini, GPT-5, and Claude. Models first assign scores independently and then engage in a debate phase, reviewing and commenting on one another’s assessments. In cases of disagreement, models may revise their scores; otherwise, initial scores are retained. In addition, models are prompted to provide rationales for their scores; however, incorporating these explanations does not improve alignment with human judgments.

The correlation between the final debate-based scores and the human ratings is measured for both human-authored and model-generated texts. As reported in Table~\ref{tab:human_model_correlation}, results show no clear improvement over prior evaluation settings. Consistent with previous findings \citep{choi2025debatevoteyieldsbetter}, multi-agent debate performs similarly to majority voting, with only negligible differences. The debate prompt template and the corresponding evaluation procedure are provided in Appendix~\ref{appendix:debate}.

\subsection{Limitations of LLM-based Evaluation}
Across the evaluated settings, LLM--human agreement varies substantially across dimensions. Alignment is consistently weaker for emotion and attractiveness than for the structured TTCT-derived dimensions, suggesting that the evaluated LLMs are less reliable for these subjective properties in our setting. This pattern is consistent with recent evidence that LLM--human agreement depends on the type of evidence required by an evaluation criterion \citep{lyu2026largelanguagemodelshumans}.

As shown in Table~\ref{tab:human_model_correlation}, majority-vote aggregation achieves broadly comparable alignment to the strongest single-model evaluations, but at increased computational cost. An alternative strategy is to assign each model to the dimension for which it demonstrates the strongest performance. Based on observed relative strengths, DeepSeek is best suited for originality, while Claude performs better on fluency and elaboration.

A more fundamental limitation lies in the sensitivity of LLM-based evaluation to prompt design. In the primary setup, each model is queried separately for its strongest dimension, for example originality for DeepSeek and fluency and elaboration for Claude. When the same questions are instead presented jointly or their phrasing is paraphrased, the resulting evaluations differ substantially despite unchanged underlying criteria. 

Table~\ref{tab:ttct_models} reports agreement under these prompting variations. The effect is particularly pronounced for \emph{fluency} and \emph{elaboration}, demonstrating that even the relative ordering assigned by an evaluator can depend on prompt formulation. This observation is consistent with recent cross-domain evidence of prompt sensitivity in LLM-based creativity evaluation \citep{lu-etal-2026-rethinking} and shows that the phenomenon also arises in short Persian literary text.

\begin{table}[t]
\centering
\large
\begin{adjustbox}{width=\columnwidth}
\setlength{\tabcolsep}{12pt}
\begin{tabular}{lccc}
\toprule
\textbf{} & \textbf{Originality} & \textbf{Fluency} & \textbf{Elaboration} \\
\midrule
Joint & 0.67 & 0.60 & 0.54 \\
Paraphrased & 0.64 & 0.33 & 0.36 \\
\bottomrule
\end{tabular}
\end{adjustbox}
\caption{Spearman correlation of model-assigned scores between the separate-question baseline and two alternative prompting settings: joint question presentation and paraphrased input. \textbf{Originality} is scored by \textit{DeepSeek-V3}, while \textbf{Fluency} and \textbf{Elaboration} are scored by \textit{Claude 3.7 Sonnet}, the best-performing model on these dimensions.}
\label{tab:ttct_models}
\end{table}

\section{\textsc{CLIN} Framework}



Motivated by the dimension-dependent behavior observed in LLM evaluation, we investigate whether the structured TTCT-derived dimensions can be approximated using dedicated, interpretable proxies. \textsc{CLIN} evaluates \emph{Originality}, \emph{Fluency}, and \emph{Elaboration} separately rather than learning a holistic creativity judge or combining the dimensions into a single aggregate score.

Specifically, \emph{originality} is approximated through global and topic-relative novelty, \emph{fluency} through contextual lexical clustering, and \emph{elaboration} through lexical diversity. This formulation differs from approaches that learn holistic evaluators from preferences or construct task-general measures of model creativity: CLIN instead associates each predefined human-rated dimension with a transparent measurement of an individual text. We additionally introduce a quality safeguard for identifying linguistically degraded outputs.

\subsection{Originality (Novelty)}

Originality reflects the degree to which a text is uncommon or unexpected relative to existing language use. In the creativity literature, it is a core component of creativity and is typically defined in terms of novelty with respect to a reference set \citep{Boden2007CreativityIA}. Prior work distinguishes between novelty relative to the full body of cultural production and novelty relative to the prior outputs of a specific individual or system \citep{Boden2007CreativityIA, franceschelli2023creativity}, differing primarily in the scope of comparison.

Following this distinction, originality is decomposed into two complementary components. \emph{Global originality} measures rarity in the language at large, while \emph{local originality} captures novelty relative to a constrained, topic-specific distribution. Dedicated metrics are defined for each component and combined into a unified score, enabling the evaluation of both globally rare expressions and context-dependent novelty. This formulation grounds originality in probabilistic notions of surprise while remaining consistent with established theories of creativity.

\paragraph{Perplexity.}

Perplexity (PPL) quantifies the statistical unpredictability of a text under a probability model \citep{Jelinek1977PerplexityaMO}; higher PPL indicates less predictable and potentially more original content. \citet{ismayilzada-etal-2025-creative} previously used PPL as a measure of surprise in preference optimization.

In this work, normalized perplexity is used as a proxy for \textit{global originality}, yielding an objective score in the range $[0,1]$ for evaluating originality in text. The full formulation and normalization procedure are described in Appendix~\ref{appendix:ppl}.

\paragraph{Diversity.}

Diversity measures how much a sentence deviates from other texts within the same topic. Sentence embeddings are computed using a multilingual sentence-transformer model\footnote{\url{https://huggingface.co/heydariAI/persian-embeddings}}, and cosine distance from the topic centroid is used as a normalized $[0,1]$ diversity score.

Perplexity and diversity are combined into a unified novelty score:
\[
\text{\textbf{Novelty}} = \text{PPL} \times \text{Diversity}
\]
where PPL denotes normalized perplexity, capturing global rarity, while Diversity captures topic-specific uniqueness. Their product therefore combines both global and local aspects of originality into a single novelty score.

\subsection{Fluency (Lexical Idea)}

In the TTCT framework, fluency reflects the ability to generate multiple relevant ideas in response to a prompt. Since explicit identification of ``ideas'' in free text is not directly observable, we propose a proxy notion of \emph{lexical ideas} to approximate idea production at the semantic level.

Unlike surface-level lexical counting, which treats each word independently, our goal is to capture semantically related words as belonging to the same underlying idea. To this end, each token is mapped into a contextual embedding space using a pretrained language model, where semantically related words are expected to have similar representations.

Formally, given a text $T$, each token $w_i \in T$ is represented as an embedding $\mathbf{h}_i$. We then apply DBSCAN clustering with cosine distance to these contextual token embeddings:
\begin{equation}
\mathcal{C}(T) =
\text{DBSCAN}(\{\mathbf{h}_1,\dots,\mathbf{h}_m\}; d_{\cos}).
\end{equation}

DBSCAN groups tokens whose contextual embeddings are sufficiently close in the embedding space, thereby forming clusters of semantically related words. We use DBSCAN because the number of semantic groups may vary across texts and is not known in advance. Unlike clustering methods that require the number of clusters to be specified beforehand, DBSCAN can determine the number of clusters directly from the density structure of the embeddings and can also identify isolated points as noise.

Fluency is then defined as the number of clusters, excluding noise. Each resulting cluster is treated as a proxy for a distinct lexical idea, providing a more meaning-aware approximation of idea production than purely surface-level lexical counting, while allowing the number of inferred ideas to vary naturally across texts.

\subsection{Elaboration (N-gram Diversity)}

To approximate the TTCT elaboration component, we use a simplified unigram diversity proxy. Elaboration is measured as the number of unique content-bearing tokens after normalization and stopword removal.

Formally, for a token sequence $X = \{x_1, x_2, \dots, x_n\}$, we compute the number of distinct filtered tokens as
$|\{x_i \in X \mid x_i \notin \text{stopwords}\}|$.
This captures lexical diversity and serves as a proxy for elaboration.



\subsection{Quality}

To ensure linguistic well-formedness and prevent reward hacking, we introduce a quality metric that captures coherence and readability, distinguishing meaningful texts from degenerate sequences that may score highly on creativity-related measures.

Quality is estimated using a contrastive perplexity-based approach that compares a sentence to minimally corrupted variants. Well-formed sentences are expected to be more predictable than their perturbed counterparts, whereas incoherent text exhibits smaller differences.

This formulation is inspired by prior work on contrastive linguistic evaluation, where language models are tested on whether they assign higher probability to well-formed sentences than to minimally different unacceptable variants \citep{marvin2018targeted, warstadt2020blimp}. Full details and validation experiments are provided in Appendix~\ref{appendix:quality}.


The metric is not used in the main analyses, as the dataset contains well-formed texts, but serves as a safeguard evaluated under controlled noise.

\begin{table}[t]
\centering
\resizebox{\columnwidth}{!}{%
\begin{tabular}{lccc}
\toprule
 & \bf Org:Novelty & \bf Flu:Lexical Idea & \bf Ela:N-gram \\
\midrule
\textbf{Correlation} & 0.45 & 0.46 & 0.67 \\
\bottomrule
\end{tabular}%
}
\caption{Spearman correlation between human-rated TTCT dimensions (\textbf{Originality, Fluency, Elaboration}) and their corresponding proxy measures (\textbf{Novelty, Lexical Idea, N-gram}), averaged over all 200 texts.}
\label{tab:ttct_human}
\end{table}

\begin{table}[t]
\centering
\small
\renewcommand{\arraystretch}{1.15}
\begin{tabular}{p{0.96\columnwidth}}
\toprule

\textbf{Sample 22} \\
\textit{Deep in my heart, melodies of longing rise, revealing my loneliness.
In these silent moments, I think of the desire to see you and write our
cherished memories in the embrace of longing.}

\vspace{2pt}
\begin{tabular}{@{}lll@{}}
\textbf{Org:Novelty} & \textbf{Flu:Lexical Idea} & \textbf{Ela:N-gram} \\
1.8 / 0.012          & 2 / 6                     & 2.6 / 17
\end{tabular}
\\
\midrule

\textbf{Sample 23} \\
\textit{When I was with you, the world was filled with the color and scent
of love. Now that you're gone, I carry the longing for you in my heart
every moment.}

\vspace{2pt}
\begin{tabular}{@{}lll@{}}
\textbf{Org:Novelty} & \textbf{Flu:Lexical Idea} & \textbf{Ela:N-gram} \\
1.2 / 0.008          & 2 / 6                     & 2.4 / 16
\end{tabular}
\\

\bottomrule
\end{tabular}

\caption{Qualitative comparison of average human-assigned TTCT scores and the corresponding \textsc{CLIN} proxy scores for two illustrative samples. Values are reported as human score / proxy score.}
\label{tab:qualitative_examples}
\end{table}

\subsection{Validating the Individual Proxies}

Following the definition of the \textsc{CLIN} proxies, we evaluate whether each proxy captures the relative differences reflected in human judgments on its corresponding TTCT dimension. Since our goal is to assess rank-order agreement rather than absolute score agreement, we use Spearman's rank correlation.

As shown in Table~\ref{tab:ttct_human}, all three proxies exhibit positive and statistically significant correlations ($p < 0.05$) with their corresponding human-rated dimensions, indicating that they preserve the ranking structure induced by human evaluations. These results provide empirical support for the use of the proposed metrics as proxies for TTCT-based creativity evaluation.

We further compare the \textsc{CLIN} proxies with Claude~3.7~Sonnet, the strongest overall zero-shot LLM judge in Figure~\ref{fig:combined_corr}, on human-authored texts. A paired item-level bootstrap comparison shows that \textsc{CLIN} significantly outperforms Claude on elaboration, while the differences for originality and fluency are not statistically significant. Thus, \textsc{CLIN} achieves comparable performance on originality and fluency and significantly better performance on elaboration. Details are provided in Appendix~\ref{appendix:bootstrap_comparison}.

To provide a qualitative illustration, Table~\ref{tab:qualitative_examples} presents two generated samples together with their human-assigned scores and corresponding proxy values. Sample 22 receives higher human scores for originality and elaboration than Sample 23, and the corresponding Novelty and N-gram values follow the same ordering. This example illustrates how the proposed proxies capture relative differences that are also reflected in human judgments.

\section{Conclusion and Future Work}

We investigated creativity evaluation in short Persian literary text from two complementary perspectives. First, we examined LLM-based evaluation across multiple models, prompting strategies, and coordination mechanisms. LLM--human agreement was strongly dimension dependent: the structured TTCT-derived dimensions of \emph{Originality}, \emph{Fluency}, and \emph{Elaboration} showed substantially stronger alignment than \emph{Emotion} and \emph{Attractiveness}, while increasingly elaborate judging procedures did not consistently improve performance and judgments remained sensitive to prompt formulation.

Second, we asked whether these structured dimensions could be evaluated without a generative judge. \textsc{CLIN} associates each dimension with a separate, interpretable proxy based on lexical, semantic, or statistical properties of the text. Despite their simplicity, the proposed proxies achieve human alignment comparable to or better than the strongest zero-shot LLM judge in our setting, while requiring substantially lower evaluation cost.

The results support a dimension-specific approach to creativity evaluation, where some human-rated properties can be approximated with simple, transparent measurements. Future work should test these proxies across languages, longer texts, and other creative tasks, and extend them to more subjective dimensions such as \emph{Emotion} and \emph{Attractiveness}.

\subsection*{Limitations}
Despite its contributions, this study has several limitations. First, the analysis is restricted to short literary text, and further experiments are needed to assess how well the findings generalize to longer and more structured creative writing tasks such as stories. Second, while the proposed proxy metrics capture key TTCT-inspired dimensions, they provide only an approximation of creativity and do not fully capture richer subjective aspects such as emotional depth and aesthetic attractiveness. Third, due to resource constraints, certain state-of-the-art large language models were not included. Evaluating the framework with a broader range of evaluator models could provide further insights into its robustness across models. Finally, as this study focuses exclusively on Persian, a low-resource language, extending it to other languages is necessary to assess cross-linguistic robustness and potential language-specific effects.

\bibliography{citations}

\appendix

\section{Appendix}

\subsection{Rubric Questions}
All creativity dimensions and their descriptions are listed in Table \ref{tab:appendix_metrics}.

\subsection{Annotator Agreement} \label{appendix:annotator_singlecol}
Given the inherently subjective nature of the evaluation task, individual annotators may differ in their judgments and assigned scores. We therefore aggregate the annotators' ratings by averaging their scores to obtain a more stable estimate of the overall human judgment. To assess the reliability of this aggregated evaluation, we compute the two-way random-effects, average-measure intraclass correlation coefficient (ICC(2,{k})), where $k=5$ corresponds to the five annotators. This measure assesses the consistency of the average ratings across annotators and provides an estimate of the reliability of the aggregated human score.


\begin{table}[H]
\centering
\setlength{\tabcolsep}{3.5pt}
\renewcommand{\arraystretch}{1.35}

{\Large
\resizebox{\columnwidth}{!}{%
\begin{tabular}{lcccccc}
\toprule
\textbf{Measure} & \textbf{Originality} & \textbf{Fluency} & \textbf{Emotion} &
\textbf{Elaboration} & \textbf{Attractiveness} & \textbf{Creativity} \\
\midrule
ICC(2,5) & 0.395 & 0.491 & 0.415 & 0.557 & 0.399 & 0.489 \\
\bottomrule
\end{tabular}%
}
}

\caption{Average inter-rater reliability across human-written and model-generated texts, measured using the two-way random-effects, average-measure intraclass correlation coefficient (ICC(2,5)).}
\label{tab:icc_average}
\end{table}

\renewcommand{\thetable}{A.\arabic{table}}
\setcounter{table}{0}

\renewcommand{\arraystretch}{1.3}  

\begin{table*}[h]
\centering
\small
\begin{tabular}{p{3.5cm} p{0.65\textwidth}}
\hline
\textbf{Rubric} & \textbf{Description} \\
\hline
Originality & To what degree does the text exhibit novel, uncommon, or unexpected ideas and expressions relative to typical responses? \\
\hline
Fluency & To what extent does the text generate a large number of different ideas? \\
\hline
Elaboration & To what extent does the text develop ideas in detail by adding explanations, descriptions, or supporting elements? \\
\hline
Emotion & How strongly and vividly does the text convey emotions and affective states to the reader? \\
\hline
Attractiveness & How appealing, engaging, or aesthetically pleasing is the text when read as a whole? \\
\hline
Creativity (Overall) & To what extent can the text be considered creative? \\
\hline
\end{tabular}
\caption{Creativity-related evaluation rubrics and guiding questions}
\label{tab:appendix_metrics}
\end{table*}

\subsection{Zero-Shot Prompt Format} \label{appendix:prompt}

The following prompt was used to evaluate Persian sentences in a zero-shot setting, based on the revised rubrics addressing construct clarity and discriminant validity. Evaluators were instructed to assign numeric scores only, according to the criteria specified.

\begin{enumerate}[label=\textbf{\arabic*.}, leftmargin=0.5cm, itemsep=1em]

\item \textbf{Originality:} The sentence's avoidance of clichés and use of unexpected word combinations.
\begin{itemize}
    \item 1: Common phrase/idiom. Contains a recognizable cliché or highly predictable word pair. No surprise.
    \item 2: Minor twist on common phrase. Modifies a familiar expression or uses a predictable but not trivial metaphor. Slightly fresh but still familiar.
    \item 3: Novel combination. No identifiable cliché. Uses an unexpected predicate or attribute. Genuinely surprising.
\end{itemize}

\item \textbf{Fluency (Idea Density):} The number of distinct concepts or clauses integrated smoothly into the sentence.
\begin{itemize}
    \item 1: One simple idea. Single clause with one subject-verb-object. No additional modifiers or subordinate clauses that add new information.
    \item 2: Two ideas. Two clauses (e.g., one main + one subordinate) or two distinct actions/attributes. Ideas are connected but not crowded.
    \item 3: Three or more ideas. Three or more clauses, or a single clause with three distinct informational elements. Still grammatically coherent.
\end{itemize}

\item \textbf{Elaboration (Sensory \& Concrete Detail):} Use of specific, sensory, or concrete details that ground abstract ideas.
\begin{itemize}
    \item 1: Weak / abstract. Uses only generic adjectives or abstract nouns. No sensory words. Feels vague.
    \item 2: Moderate. One or two specific details (e.g., a concrete object name, a smell, a sound). Some sensory input but not sustained.
    \item 3: High / vivid. Uses unexpected concrete details across multiple sensory modalities (sight, sound, touch, taste, smell). Descriptions are precise, surprising, or highly evocative.
\end{itemize}

\item \textbf{Emotion:} The sentence's ability to evoke a specific, discrete emotion in the reader.
\begin{itemize}
    \item 1: No specific emotion. Neutral, clinical, or purely factual. No emotional vocabulary or evocative imagery.
    \item 2: Recognizable but mild. Clearly aims for an emotion (e.g., sadness, joy, fear, nostalgia) but feels muted or generic. Present but not impactful.
    \item 3: Highly emotional and impactful. Emotion is unmistakable and strong. Uses visceral imagery, rhythm, or word choice that creates an emotional response.
\end{itemize}

\item \textbf{Attractiveness (Aesthetic Coherence):} The internal harmony between word choice, rhythm, imagery, and tone.
\begin{itemize}
    \item 1: Jarring or flat. Words clash in tone (e.g., mixing very formal and very colloquial language without purpose). Rhythm is awkward. No sense of style.
    \item 2: Mostly coherent. Tone is consistent. No major clashes. Some rhythm or imagery, but feels ordinary or slightly bland. Readable but not beautiful.
    \item 3: Highly harmonious. Word sounds, sentence length, and imagery work together. May use prosody, alliteration, parallelism, or a surprising but fitting metaphor. Feels purposeful and pleasing.
\end{itemize}

\item \textbf{Creativity (Overall):} The sentence as a whole feels novel, surprising, and valuable – an emergent effect beyond the sum of its parts.
\begin{itemize}
    \item 1: Not creative at all. Predictable, generic, or assembled from common templates. No surprise. Could have been written by any novice.
    \item 2: Somewhat creative. Contains one unexpected element (e.g., a fresh word choice or a small twist) but overall structure is conventional. Mild surprise, limited lasting impression.
    \item 3: Fully creative. Evokes “I wouldn’t have thought of that.” Multiple elements interact to produce surprise, and the result feels meaningful, vivid, or insightful. Memorable.
\end{itemize}

\end{enumerate}

\medskip

\noindent
\textbf{Important:} Only provide numeric scores for each criterion (1, 2, or 3). Do not provide explanations.  

\begin{verbatim}
Originality: (number)
Fluency: (number)
Elaboration: (number)
Emotion: (number)
Attractiveness: (number)
Creativity: (number)
\end{verbatim}

\subsection{Multi-Agent Debate Template} \label{appendix:debate}

\begin{figure}[H]
    \centering
    \includegraphics[width=0.7\columnwidth, height=8cm]{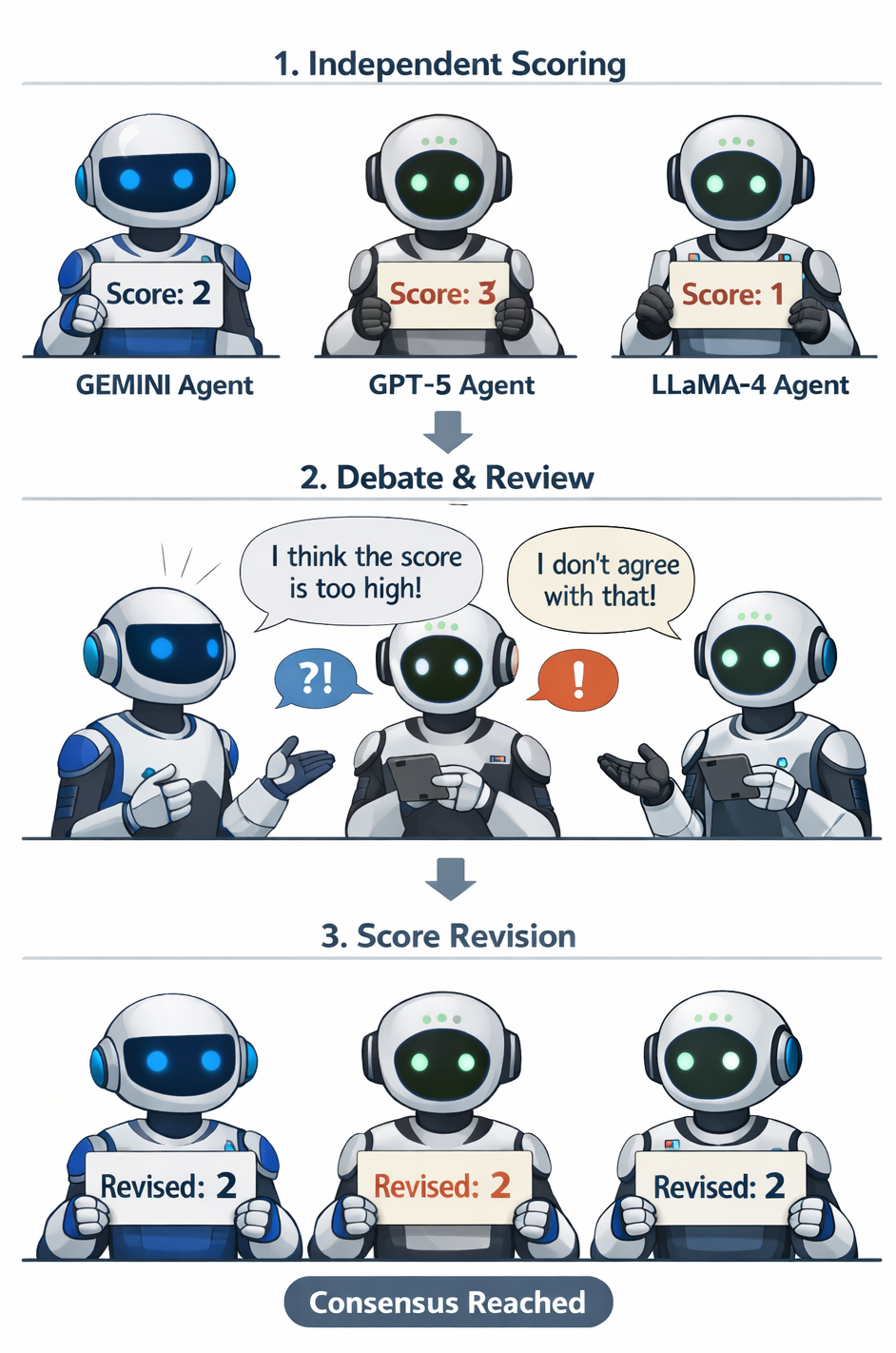} 
    \caption{Multi-agent debate process.}
    
\end{figure}

For each model participating in the multi-agent debate, the prompt is structured as follows:

\begin{quote}
\textbf{Evaluation Question:} \textit{\{evaluation\_question[dimension]\}}\\[1em]
Here are the answers from other models in the previous round: \\
\texttt{\{previous\_info\}}\\[0.5em]
Your previous answer for this sentence: \\
\texttt{\{last\_answers[model]\}}\\[0.5em]
If you consider it necessary, revise your score based on the previous round's feedback and other models' answers.
\end{quote}

\noindent In this setup, each model first evaluates each sentence independently. In the next round, it observes the previous responses of other agents along with its own prior score and is asked to update or confirm its evaluation, enabling a structured multi-agent discussion that promotes alignment and consistency. If the models disagree in the final round of the multi-agent debate on whether an update should be applied, we determine the final decision using majority voting among the participating models.

\subsection{Normalized Perplexity} \label{appendix:ppl}
The per-token negative log-likelihood (NLL) of each sentence is computed under a pretrained language model and converted to perplexity (PPL), which quantifies how well the model predicts a sequence. Formally, perplexity is defined as:

\begin{equation}
\text{PPL}(x) =
\exp\left(
-\frac{1}{N} \sum_{i=1}^{N} \log P(w_i \mid w_{<i})
\right),
\end{equation}

\noindent where $P(w_i \mid w_{<i})$ denotes the probability assigned to token $w_i$ given its preceding context.

Since PPL is unbounded and scale-dependent, it is mapped to the fixed range $[0,1]$ using a clipped min--max normalization. Specifically, given predefined bounds $\text{PPL}_{\min}$ and $\text{PPL}_{\max}$, the raw perplexity is first clipped:

\begin{equation}
\tilde{\text{PPL}}(x) =
\mathrm{clip}(\text{PPL}(x), \text{PPL}_{\min}, \text{PPL}_{\max}),
\end{equation}

\noindent and then normalized as:

\begin{equation}
\text{Score}(x) =
\frac{\tilde{\text{PPL}}(x) - \text{PPL}_{\min}}
{\text{PPL}_{\max} - \text{PPL}_{\min}}.
\end{equation}

In all experiments, we set $\text{PPL}_{\min} = 1.0$ and $\text{PPL}_{\max} = 1000.0$. This transformation assigns lower scores to highly probable (i.e., conventional) sentences and higher scores to less likely ones, while ensuring numerical stability and robustness to outliers.

\subsection{Quality Metric under Controlled Corruption} \label{appendix:quality}

Quality is computed using a contrastive perplexity-based formulation. Given a sentence $s$, its perplexity $\text{PPL}(s)$ is calculated using a language model. We generate minimally corrupted variants by applying token deletion and local shuffling, and compute their average perplexity $\overline{\text{PPL}}_{\text{cor}}$. The quality score is then defined as:

\begin{equation}
    r = \frac{\overline{\text{PPL}}_{\text{cor}}}{\text{PPL}(s)}, \qquad
    \text{Quality}(s) = \frac{r}{1 + r}.
\end{equation}

This formulation captures the relative gap between a sentence and its corrupted variants. Well-formed sentences are expected to have substantially lower perplexity than their perturbed counterparts, resulting in higher quality scores. In contrast, incoherent or degenerate sequences tend to exhibit smaller differences, yielding lower scores.

To validate the sensitivity of the metric, we conduct a controlled perturbation experiment in which a fixed proportion of tokens in each sentence is randomly shuffled. The shuffle ratio is varied from 0.0 to 0.5, progressively degrading coherence while preserving the underlying token distribution. For each corruption level, quality scores are computed and averaged across the dataset.

As shown in Table~\ref{tab:quality_perturbation}, the metric exhibits a consistent monotonic decrease as the degree of perturbation increases, indicating that it effectively captures disruptions in linguistic well-formedness. This behavior supports its use as a robust proxy for textual quality.

\begin{table}[H]
\centering
\small
\begin{tabular}{lccccc}
\toprule
Shuffle Ratio & 0.0 & 0.1 & 0.2 & 0.3 & 0.5 \\
\midrule
Quality Score & 0.848 & 0.821 & 0.763 & 0.699 & 0.614 \\
\bottomrule
\end{tabular}
\caption{Effect of controlled token shuffling on the proposed quality metric. Increasing the proportion of shuffled tokens leads to a consistent degradation in quality scores, demonstrating sensitivity to disruptions in coherence and well-formedness.}
\label{tab:quality_perturbation}
\end{table}

\subsection{Reference-based Evaluation: Experimental Details and Results}
\label{appendix:comparative}

This section provides detailed descriptions and results for the reference-based evaluation setting. We evaluate three of the best-performing models from the main experiments, namely Claude~3.7~Sonnet, DeepSeek-V3, and GPT-5, across all six dimensions.

In this setup, evaluation is formulated as a pairwise comparison task, where the model selects the better text between a candidate and a reference for a given dimension. For each run, reference sentences are randomly sampled from texts within the same topic in the dataset, ensuring topical consistency. No sentence is compared against itself.

To enable comparison with human judgments, we convert absolute human annotations into pairwise preferences, producing comparative human scores. Model decisions are then evaluated by computing correlation with these human-derived comparative judgments.

Results are averaged over three runs with different randomly sampled references and further aggregated across both human-authored and model-generated texts. The final results are summarized in Table~\ref{tab:comparative_correlation}.

\begin{table}[]
\centering
\setlength{\tabcolsep}{5pt}
\renewcommand{\arraystretch}{1.15}
\scalebox{0.8}{
\begin{tabular}{lcccccc}
\toprule
\textbf{Model} & \textbf{Org.} & \textbf{Flu.} & \textbf{Ela.} & \textbf{Emo.} & \textbf{Crea.} & \textbf{Attr.} \\
\midrule
Claude~3.7~Sonnet & 0.28 & 0.21 & 0.57 & 0.12 & 0.18 & 0.34 \\
DeepSeek-V3 & 0.18 & 0.31 & 0.36 & 0.16 & 0.13 & 0.14 \\
GPT-5 & 0.31 & 0.39 & 0.55 & 0.21 & 0.19 & 0.20 \\
\bottomrule
\end{tabular}
}
\caption{Average correlation between model pairwise judgments and human comparative preferences across six dimensions. Results are averaged over three runs with different randomly sampled references and across both human-authored and model-generated texts.}
\label{tab:comparative_correlation}
\end{table}

\begin{table}[]
\centering
\setlength{\tabcolsep}{10pt}
\renewcommand{\arraystretch}{1.15}
\scalebox{0.9}{
\begin{tabular}{lcc}
\toprule
\textbf{Dimension} &
\textbf{95\% CI for $\Delta\rho$} &
\textbf{$p$-value} \\
\midrule
Originality
    & $[-0.1097,\,0.2976]$
    & 0.3606 \\
Fluency
    & $[-0.1960,\,0.2467]$
    & 0.8313 \\
Elaboration
    & $\mathbf{[0.0536,\,0.4358]}$
    & \textbf{0.0111} \\
\bottomrule
\end{tabular}
}
\caption{Paired item-level bootstrap comparison between the dimension-specific \textsc{CLIN} components and Claude~3.7~Sonnet on human-authored texts. Differences are defined as $\Delta\rho=\rho_{\textsc{CLIN}}-\rho_{\text{Claude}}$, such that positive values favor \textsc{CLIN}. Boldface indicates a statistically significant difference at $p<0.05$.}
\label{tab:bootstrap_clin_claude}
\end{table}

\subsection{Paired Bootstrap Comparison with the Best Zero-shot LLM Judge}
\label{appendix:bootstrap_comparison}

We compare the dimension-specific \textsc{CLIN} components with Claude~3.7~Sonnet, the strongest overall zero-shot LLM judge, using a paired item-level bootstrap on the human-authored texts. At each iteration, items are sampled with replacement while preserving the paired human rating, \textsc{CLIN} score, and Claude score for each item. We then compute the Spearman correlation of each evaluator with human judgments and record their difference, defined as $\Delta\rho=\rho_{\textsc{CLIN}}-\rho_{\text{Claude}}$. The resulting bootstrap distribution is used to estimate the 95\% confidence interval and the $p$-value for the null hypothesis that $\Delta\rho=0$.

As shown in Table~\ref{tab:bootstrap_clin_claude}, \textsc{CLIN} significantly outperforms Claude~3.7~Sonnet on elaboration, with a 95\% CI of [0.0536, 0.4358] ($p=0.0111$). For originality and fluency, the differences are not statistically significant. These results indicate that \textsc{CLIN} performs comparably to the strongest zero-shot LLM judge on originality and fluency, while providing significantly better performance on elaboration.

\subsection{Human Annotators}
The annotations were performed by graduate students who are native speakers of Persian. After an initial calibration session to ensure a shared understanding of the evaluation criteria, annotators completed the task independently. Annotators were compensated on an hourly basis at a rate aligned with local fair compensation standards, and the annotation process took approximately two hours per annotator. The annotation guidelines are publicly available in the project \href{https://github.com/smrmodares/CLiN}{GitHub} repository.

\subsection{Intended Use of Data Artifacts}
We use publicly available datasets \citep{tourajmehr-etal-2025-evaluating} released for research purposes and confirm that our use complies with their intended use and licensing conditions. We release our annotated derivatives under the ACL Community Agreement for the use of materials, maintaining the same research-only usage constraints.

\end{document}